\documentclass[10pt]{article}
\usepackage{amsmath}
\usepackage{amsthm}
\usepackage{verbatim}
\usepackage{latexsym}
\usepackage{amssymb}
\usepackage{amscd}

\usepackage[dvips]{graphicx}
\usepackage{times}

\input epsf
\theoremstyle{plain}
\newtheorem{thm}{Theorem}[section]

\newtheorem{lemma}[thm]{Lemma}

\theoremstyle{definition}

\newtheorem{definition}[thm]{Definition}

\theoremstyle{remark}

\def\mylabel#1{\label{#1}}

\newcommand{\Figure}[3]{
\begin{figure}[htbp]\begin{center}{#1}\caption{#2}\label{#3}\end{center}\end{figure}
}

\begin{document}
\title{Structure from Motion: Theoretical Foundations of a Novel Approach Using Custom Built Invariants\thanks{This work was supported by NSF grants KDI BCS-9980091 and 0074276.}
}

\author{\begin{tabular}[t]{c@{\extracolsep{8em}}c}
Pierre-Louis Bazin${}^1$   &   Mireille Boutin${}^2$  \\
\end{tabular}\vspace{0.5cm} \\
 ${}^1$ Department of Engineering, \\
 ${}^2$ Division of Applied Mathematics, \\
 Brown University \\
 Providence RI 02912, USA.\\
}
\date{}
\maketitle

\begin{abstract}
We rephrase the problem of 3D reconstruction from images
in terms of intersections of projections of orbits of custom built Lie groups actions.
We then use an algorithmic method based on moving frames {\em \`a la Fels-Olver}
to  obtain a fundamental set of invariants of these  groups actions.
The invariants are used to define a set of equations to be solved by the points of the 3D object, providing a new technique for recovering 3D structure from motion.
\end{abstract}

\section{Introduction}

The concept of invariance is of major importance in modern geometry. 
In the field of computer vision, 
invariants have been  used for more than a decade 
for object recognition and reconstruction 
(see for example \cite{mundy92,mundy94}).

One particular vision problem one could hope to solve using invariants is that of reconstructing a 3D object from a set of pictures taken from unknown viewpoints.
For example, given $n$ ordered points 
${\mathcal O}_1, {\mathcal O}_2,\ldots ,{\mathcal O}_n \in {\mathbb R}^3$ (the object) and $t$ sets of $n$ ordered points $p_1,\ldots, p_n \in {\mathbb R}^2$ (for each of the pictures), we would like to determine the object in ${\mathbb R}^3$ from the images. 

One possible way of solving this problem would be to
define an equivalence relation between 
all the possible pictures of an object 
and to find functions computed on the pictures that are constant on each equivalence class.
Characteristics of the object could be inferred from these functions, regardless of the camera position relative to the object.
To define such an equivalence class, 
one can try to use the orbits of a group action. Real valued functions that are constant on the orbits of a group action are called invariants.
We would thus be interested in finding invariants of a group action which would be transitive on the set of pictures of any given object, i.e.~``view invariants''.

Unfortunately, as is commonly known in the vision community, 
{\em view invariants do not exist for 3D point sets 
of arbitrary size (in general position)}. One can still build invariants for specific objects (for instance, planar sets of points, pencils of lines, etc.) but not for arbitrary shapes.
The problem is that the set of pictures of an object intersects with the set of pictures of other objects. 
Observe that if a view invariant $I$ takes a constant value $c$ for all pictures of an object $O$, 
then $I$ is also equal to $c$ on the set of pictures of any object whose set of pictures intersect with the set of pictures of $O$.  
One can actually show \cite{burns92} that any equivalence relation between all the pictures of each objects defines a unique equivalence class on the space of pictures and thus, any view invariant is trivial.
From a group point of view, 
this means that any group action that is transitive 
on the set of pictures of any object must be transitive on the set of {\em all} pictures.

Another way to solve this problem would be to use the reverse approach: 
try to characterize all the possible objects corresponding to a given picture.
It would be useful to find  functions 
that are constant on equivalence classes 
that include all the objects  corresponding to a given picture.
However,  it is easy to see that such equivalence classes of objects are in one to one correspondence with the equivalence classes of pictures discussed earlier, and thus that only one such equivalence class can exist. There are therefore no ``object invariants''.
Nevertheless, given a view of an object, we {\em can} infer information about the object, so there must be a way to overcome this difficulty.

It is the fact that we know how a camera builds images (using the perspective projection) that will allow us to   build an equivalence relation 
which characterizes objects corresponding to different pictures. 
The key is to consider an equivalence relation on a higher dimensional space,  sort of {lifting} the set of objects of different pictures to different {\em ``heights''} in the extra dimensions. 
We can do this 
using the three extra dimensions provided by the camera center position.
More precisely, 
we can construct a Lie group action 
on the object points ${\mathcal O}_1,\ldots , {\mathcal O}_n$ and the camera center $P_0$ 
which summarize what is unknown about the object-camera system given a picture of an object.
Invariants of this group actions prove to be sufficient for solving the  problem of recovering the object coordinates in ${\mathbb R}^3$, i.e. the "structure from motion" problem.

Recent advances in the theory of moving frames \cite{FO1,FO2}
provide us with a systematic way to proceed to obtain the invariants 
of any given (regular) Lie group action. 
Using this systematic method, 
we obtain the invariants of our custom built action without any difficulty. 
We can even venture a bit further 
by modifying the group action and repeat the exercise so to obtain even 
more new tools for 3D structure and motion analysis.

We begin our exposition with a summary of some of the relevant theoretical aspects of the theory of moving frames  and its  application to the computation of (joint) invariants.  
In Section 3, we formulate the problem of structure from motion in this framework and obtain the corresponding invariants.
We end the section by remarking how this leads to a simple test
to identify camera motions that are pure rotations.
Section 4 presents some variations of our approach 
where more camera parameters are considered, 
including a generalization to the case of a variable focal distance.

\section{Theoretical Foundations}

Let $M$ be an $m$-dimensional smooth (Hausdorff) manifold and $G$ be an $r$-dimensional Lie group acting (smoothly) on $M$.

\begin{definition}
An {\em invariant} is a function $I: M\rightarrow {\mathbb R}$ which remains unchanged under the action of the group. In other words, 
\[I(g\cdot z)= I(z), \text{ for all }z\in M \text{ and all }g\in G.
\]
A {\em local invariant is } a function $I: M\rightarrow {\mathbb R}$ for which there exists $N_e$, a neighborhood of the identity $e\in G$, such that
\[I(g\cdot z)= I(z), \text{ for all }z\in M \text{ and all }g\in N_e.
\]
\end{definition}

\begin{definition} 
We say that $G$ acts {\em semi-regularly} on $M$ if all orbits have the same dimension. If, in addition, any point $p_0 \in M$  is surrounded by an arbitrarily small neighborhood whose intersection with the orbit through $p_0$ is connected, then we say that $G$ acts {\em regularly}.
\end{definition}

The following theorem, due to Frobenius \cite{Frobenius}, is of central importance to our approach. A proof can be found in \cite{Obook3}.
It provides us with a simple way of characterizing the orbits using invariants.

\begin{thm}[Frobenius Theorem]
\mylabel{fundamental theorem}
If  $G$ acts on an open set $O\subset M$ semi-regularly 
 with $s$-dimensional orbits, then  $\forall p_0 \in O$ there exist $m-s$ functionally independent local invariants $I_1,\ldots,I_{m-s}$  defined on a neighborhood $U$ of $p_0$ such that any other local invariant $H$ defined near $p_0$ is a function $H=f(I_1,\ldots,I_{m-s})$. 
If $G$ acts regularly on $O$, then we can choose $I_1,\ldots,I_{m-s}$ to be  global invariants on $O$. In that case, two points $p_1, p_2 \in O$ are in the same orbit relative to $G$ if and only if $I_i (p_1)= I_i (p_2)$, for all $ i=1,\ldots ,m-s$. 
\end{thm}

By {\em functional independence} of the (smooth) functions  $I_1,\ldots,I_{m-s}$ on an open set $O$, we simply mean that the Jacobian matrix of  $I_1,\ldots,I_{m-s}$
 has maximal rank $m-s$ on an open and dense subset of $O$.
The set $\{ I_1,\ldots,I_{m-s} \}$ is often called a {\em complete fundamental set of invariants on $O$}. Note that the complete fundamental set of invariants is by no mean unique.

As we shall consider actions on points, we are interested in the case where $M={\mathcal V}\times {\mathcal V}\times \ldots \times {\mathcal V}$ ($n$-times) $=:{\mathcal V}^{\times (n)} $ is the Cartesian product of $n$ copies of a manifold ${\mathcal V}$. 

\begin{definition}
We say that $G$ acts 
{\em diagonally} on ${\mathcal V}^{\times (n)}$ if there exists an action of $G$ on ${\mathcal V}$ such that
for any $g\in G$, for any $n\in {\mathbb N}$ and any $z_1,\ldots , z_n \in {\mathcal V}$, the action $g \cdot (z_1,\ldots, z_n)$ can be written as
\[g \cdot (z_1,\ldots, z_n)=(g*z_1,\ldots, g*z_n).
\] 
\end{definition}

The group actions we will define for our object-camera systems
are not diagonal actions. However, we will find a normal subgroup $H$ 
 of $G$  such that
\begin{enumerate}
\item  The subgroup $H$ can be written as $H=\mathfrak{H} \times \ldots \times \mathfrak{H}  $ ($n$-times)
where $\mathfrak{H}$ is a Lie group acting on ${\mathcal V}$ 
in a consistent manner with the action of $H$, i.e.~$(h_1, \ldots , h_n) \cdot (z_1,\ldots, z_n)= ( h_1* z_1 ,\ldots , h_n* z_n) $, for all 
$h_1,\ldots, h_n \in {\mathfrak H}$ 
and all $z_1 ,\ldots ,z_n \in {\mathcal V}$.
This insure that 
 ${\mathcal V}^{\times (n)}/H$ can be written as 
${\mathcal V}^{\times (n)}/H=({\mathcal V}/{\mathfrak H})^{\times (n)}$.

\item The action of $G/H$ on ${\mathcal V}^{\times (n)} / H=({\mathcal V}/\mathfrak{H})^{\times (n)}$ is diagonal.
\end{enumerate}
So for all practical purposes, we shall ultimately have to deal with diagonal group actions.

In our approach to structure from motion, 
invariants are used to obtained equations 
that must be satisfied by the object and the camera. 
The more invariants we have, 
the more equations need to be satisfied. 
We need enough equations to completely determine the object. 
Observe that the dimension of the orbit is bounded 
by the dimension of the group. 
So in the case of a diagonal action,  
taking more and more copies of ${\mathcal V}$ (i.e.~ more and more points) 
allows for the existence of as many invariants as necessary.
The question that remains is: 
how can we obtain an expression for these invariants?
Thanks to recent advances in the theory of moving frames  \cite{FO1,FO2}, 
this problem can now be solved in an algorithmic fashion. 
We now summarize some of this theory
and show how moving frames can be used as a tool 
to obtain a complete set of fundamental invariants.

\begin{definition}
A {\em (right) moving frame} is a map $\rho : M\rightarrow G$ which is (right) equivariant, i.e.~$\rho (g\cdot z)=\rho(z) g^{-1}$, for all $g\in G$ and $z\in M$.
\end{definition}

Unfortunately moving frames do not exist for all group actions.

\begin{thm}
A moving frame exists if and only if the action if the group action satisfies
\[ \{g\in G | \exists z\in M, g\cdot z=z \}=e,
\]
where $e$ denotes the identity in $G$. This property is called {\em freeness } of the group action.
\end{thm}

Demanding freeness of the group action is very strong. It appears that, in order to be able to deal with the generic cases, we need to relax this condition a little bit.

\begin{definition}
A {\em local moving frame} is  a map $\rho : M\rightarrow G$ such that $\rho (g\cdot z)=\rho(z) g^{-1}$, for all $g\in N_e$, a neighborhood of the identity  $e\in G$,  and all $z\in M$.
\end{definition}

\begin{thm}
A local  moving frame exists if and only if there exists a neighborhood $N_e$ of the identity in $e\in G$ such that
\[ \{g\in N_e | \exists z\in M, g\cdot z=z \}=e,
\]
or equivalently, if and only if for all $z\in M$, 
the dimension of the orbit through $z$ 
is equal to $r$, the dimension of $G$.  
This property is called {\em local freeness } of the group action.
\end{thm}

Our hope is that, by acting diagonally on more and more copies of a manifold, we shall eventually obtain a locally free action. 
So we are interested in determining a simple criterion on the group action of $G$ on ${\mathcal V}$ to determine whether or not this will work out.

\begin{definition}
We say that $G$ acts on $M$ {\em effectively} if
\[\{g\in G |\quad g\cdot p=p, \text{ for all }p\in M \}= \{e \}.
\]
We say that $G$ acts on $M$ {\em locally effectively} if
\[\{g\in G |\quad g\cdot p=p, \text{ for all }p\in M \} \mbox{ is a discrete subgroup of $G$.}
\]
\end{definition}

Many groups do not act effectively.
However, given $G$ acting non effectively on $M$, we can consider $\tilde{G}=G/G_M$, where $G_M= \{g\in G |g\cdot z=z, \forall z\in M  \}$, which acts in essentially the same way as $G$ except that it acts effectively.
Unfortunately, effectiveness is not quite enough for achieving our goal.

\begin{definition}
We say that $G$ acts  {\em effectively on subsets of $M$} if, for any open subset $U\subset M$,
\[\{g\in G |\quad g\cdot p=p, \text{ for all }p\in U \}= \{e \}.
\]
We say that $G$ acts {\em locally effectively on subsets of $M$} if,  for any open subset $U\subset M$,
\[\{g\in G |\quad g\cdot p=p, \text{ for all }p\in M \} \mbox{ is a discrete subgroup of $G$.}
\]
\end{definition}
Observe that effectiveness on subsets implies effectiveness. The converse is not true in general, but it holds for all analytic group actions.

\begin{thm}
\mylabel{freeness}
If a group $G$ acts on a manifold ${\mathcal V}$ 
locally effectively on subsets, then for $n\in {\mathbb N}$ big enough, the  induced diagonal action of $G$ on ${\mathcal V}^{\times (n)}$ is locally free on on open and dense subset of  ${\mathcal V}^{\times (n)}$. This is equivalent to saying that the orbit dimension is  equal to the dimension of $G$ on this open and dense subset.
We denote by $n_0$ the minimal integer for which this is true.
\end{thm}

So at least for all analytic group actions 
(moding out by a subgroup if necessary), 
acting (diagonally) on more and more copies of a manifold will eventually lead to a regular and locally free action on an open subset. 
A local moving frame will thus exist and provide us with the tools we need  to obtain a complete fundamental set of invariants so completely characterize the orbits.

We now explain how to construct a (local) a moving frame and to obtain a complete fundamental set of invariants.
A more detailed exposition can be found in \cite[Chapter 8]{Obook3}.

Let $g=(g_1,\ldots, g_r)$  be local coordinates  for $G$ 
in a neighborhood of the identity. 
Suppose that $G$ acts regularly on $M$.
For simplicity, let us  assume in addition that the orbits of $G$ have the same dimension $r$ as $G$ itself. 
In other words, we are assuming that the action is locally free. 
A simple variation allowing us to deal with merely regular actions 
will also be explained shortly.

\begin{itemize}
\item{Step 1:} 
Write down the group transformation equations $\bar{x}=g\cdot x$ explicitly.
\[
\left\{
\begin{array}{ccc}
\bar{x}_1&=&f_1(g_1,\ldots ,g_r,x_1,\ldots ,x_m),\\
&\vdots&\\
\bar{x}_m&=&f_m(g_1,\ldots ,g_r,x_1,\ldots ,x_m).\\ 
\end{array}
\right.
\]
\item{Step 2:}
Choose constants $c_1,\ldots ,c_r\in{\mathbb R}$ and set $r$ of the transformed coordinates equal to those constants. For simplicity, we relabel the coordinates and write
\begin{equation}
\mylabel{normalizations}
\left\{
\begin{array}{ccc}
f_1(g_1,\ldots ,g_r,x_1,\ldots ,x_m)&=&c_1,\\
&\vdots&\\
f_r(g_1,\ldots ,g_r,x_1,\ldots ,x_m)&=&c_r.\\
\end{array}\right.
\end{equation}
These equations are called the {\em normalization equations}.
\item{Step 3:}
Solve the normalization equations  for $g=(g_1,\ldots ,g_r)$. 
The solution $ g=\rho(x)$ is a moving frame.

\item{Step 4:} Compute the action of the moving frame on the remaining coordinates.
The set of resulting functions 
\[
\left\{
\begin{array}{ccc}
\left. \bar{x}_{r+1}\right|_{g=\rho(x)}&=&I_1(x_1,\ldots ,x_m),\\
&\vdots&\\
\left. \bar{x}_{m}\right|_{g=\rho(x)}&=&I_{m-s}(x_1,\ldots ,x_m).\\
\end{array}
\right.
\]
is a complete fundamental set of local invariants.

\end{itemize}

The choice of constants in Step 2 is somewhat arbitrary: 
we are free to choose any numbers for which a solution to the normalization equations exists, provided that these constants define a cross-section (i.e.~provided that the normalization equations define a submanifold which is transversal to the orbits).
To simplify the solving process, it is usually a good idea to choose as many constants as possible to be zero.

If the action is not free but merely regular, 
we can still find a system of functionally independent local invariants. What we do is the following. 
Let $s$ be the dimension of the orbits of $G$ ($s<r$). We solve the $s$ equations $f_1(g,x)=c_1,\ldots ,f_s(g,x)=c_s$ 
for $s$ of the group parameters and replace them in the remaining equations 
$\overline{x}_{s+1}=f_{s+1}(g,x),\ldots ,\overline{x}_m=f_m(g,x)$ 
to get the $m-s$ invariants. The other group parameters will not appear 
in the equations. This procedure is called a {\em partial moving frame normalization method}.

Equipped with these theoretical tools, the computation of invariants becomes a simple systematic procedure. 
We can thus feel free to consider any Lie group action imaginable 
and try to obtain its invariants. 
As we have seen, in theory, results are guaranteed provided that the group action is locally effective on subsets, which we can always arrange in the case of analytic group actions.


\section{Application to 3D Shape Reconstruction}

Let us think for a moment about the process of taking a picture. 
This process involves, first of all, the placement of a camera in space. Then, particles of light starting from the shape (points in ${\mathbb R}^3$) travel on a straight line in the direction of the camera center, leaving its trace on a film, i.e.~on the intersection of the picture plane and the straight travel line.
So to the picture-camera system placed somewhere in ${\mathbb R}^3$, 
there corresponds a set of $n$ straight lines in ${\mathbb R}^3$ representing paths of light going from the object to the camera center.

This process can be seen as a group action generated by an action of $SE(3)$ and an action of ${\mathbb R}^n$ on the camera center together $P_0$ with the image points $P_1,\ldots, P_n$. 
The idea is to allow each $P_i$ to move independently along the line through $P_i$ and $P_0$, while allowing the line configuration to be rotated and translated in space. 
As discussed in the introduction,
including the camera center on the space acted on by the group will allow us
to obtain significant equivalence classes. 
This is the key to guarantee the existence of non-trivial invariants
and these invariants can be obtained by our systematic method.

So, 
given is a 2D image depicting $n$ points $p_1,\ldots, p_n \in {\mathbb R^2}$. 
We assume this picture was taken by a camera with fixed internal parameters. 
These parameters can be calibrated beforehand, so that the focal length is ${\mathcal F}=1$\footnote{the value is arbitrary, it simply fixes the overall scale of the 3D reconstruction.} and the 2D image coordinates match the 3D coordinates as defined below.
We embed the picture-camera system in ${\mathbb R}^3$ by setting 
the camera center to be $\tilde{p}_0=(0,0,0)$ and the picture points $\tilde{p}_i$'s to be $\tilde{p}_i=p_i\times  {\mathcal F}$. 
This is of course, in general, not 
the actual position in which the picture was taken.
However, there exists a rigid transformation $g\in SE(3)$ 
such that 
$g\cdot (\tilde{p}_0,\tilde{p}_1,\ldots, \tilde{p}_n)= ({\mathfrak P}_0, {\mathfrak P}_1,\ldots ,{\mathfrak P}_n)$ corresponds to the actual position of the picture-camera system at the moment where the picture was taken. In other words, if the object is made of $n$ points, say ${\mathcal O}_1, \ldots, {\mathcal O}_n$, 
then each transformed image point ${\mathfrak P}_i$ lies on the straight line passing through ${\mathcal O}_i$ and the  camera center ${\mathfrak P}_0$. 
In order to fully formulate the problem in terms of orbits, we want to consider smooth transformations that will map the image points 
$( {\mathfrak P}_1,\ldots ,{\mathfrak P}_n)$ to the object points 
$({\mathcal O}_1, \ldots,  {\mathcal O}_n )$. 
For this, we allow each point  $ {\mathfrak P}_i$ 
to move independently along each ray of light so to go back to its source on the object.

We would like to determine where ${\mathfrak P}_0$ 
and the ${\mathcal O}_i$'s lie. 
Given a picture, 
it is of course impossible to determine the points $({\mathfrak P}_0,{\mathcal O}_1,\ldots ,{\mathcal O}_n) $. 
However the ``line arrangement'' defined by the picture-camera system 
provides us with important information. 
In particular, we know that   
the points $({\mathfrak P}_0,{\mathcal O}_1,\ldots ,{\mathcal O}_n) $ 
lie  on the orbit 
of the Lie group action 
on $\{ (P_0,P_1,\ldots, P_n)\in {\mathbb R}^3\times ({\mathbb R}^3)^{\times (n)} \}$ defined by:
\begin{eqnarray*}
\bar{P}_0 &= & R P_0 +T \\
\bar{P}_i &= & R (P_i+\lambda_i (P_i-P_0))+T, \text{ for }i=1,\ldots, n, \\
\end{eqnarray*}
with $R\in SO(3)$ a rotation, $T\in {\mathbb R}^3$ a translation and $\lambda_i\in {\mathbb R}$, a factor of depth, for $i=1,\ldots, n$.
Observe that the action of ${\mathbb R}^n$ parameterized by the $\lambda$'s
commutes with the action of $SE(2)$ generated by the rotation $R$ and the translation $T$.
Therefore, this defines  a finite dimensional Lie group action, more precisely the action of a $(6+n)$-dimensional Lie group on a $(3n+3)$-dimensional manifold. 
Note that, although invariant-based techniques have already been used in the field for the general projective and affine transformation groups~\cite{quan95,weiss96,sparr98}, the Lie groups we shall define have little to do with these traditional transformation groups.

Assuming that the picture points are distinct, 
then the group action is regular and
the orbits are $6$-dimensional, for $n=1$, and 
$(6+n)$-dimensional as soon as $n\geq 2$. 
Therefore by Theorem \ref{fundamental theorem}, there are $2n-3$ fundamental invariants whenever $n\geq 2$.
We now follow the steps of the moving frame normalization method to obtain them.

We set 
\begin{eqnarray*}
\bar{P}_0 &= &(0,0,0)^T ,\\
\bar{P}_1 \cdot (0,1,0) &=&0,\\
\bar{P}_1 \cdot (0,0,1) &=&0,\\
\bar{P}_2 \cdot (0,0,1) &=&0,\\
\text{and } \bar{P}_i \cdot (1,0,0) &=&1, \text{ for all }i=1,\ldots,n.
\end{eqnarray*}
Solving for the group parameters, we obtain 
\begin{equation}
\label{eqn:first}
\begin{array}{rcl}
T &=& -R P_0,\\
R &=& R_1 R_2 R_3, \\
R_1&=&\left(
\begin{array}{ccc}
1 & 0 & 0 \\
0 & \frac{f}{\sqrt{f^2+g^2}} &  \frac{g}{\sqrt{f^2+g^2}} \\
0 &  -\frac{g}{\sqrt{f^2+g^2}} & \frac{f}{\sqrt{f^2+g^2}}
\end{array} 
\right), \\
R_2&=&
\left( 
\begin{array}{ccc}
\frac{\sqrt{x_1^2+y_1^2}}{\sqrt{x_1^2+y_1^2+z_1^2}} & 0 & \frac{z_1}{\sqrt{x_1^2+y_1^2+z_1^2}} \\
0 & 1 & 0 \\
\frac{-z_1}{\sqrt{x_1^2+y_1^2+z_1^2}} & 0 & \frac{\sqrt{x_1^2+y_1^2}}{\sqrt{x_1^2+y_1^2+z_1^2}}
\end{array}
\right), \\
R_3 &=&
\left( 
\begin{array}{ccc}
\frac{x_1}{\sqrt{x_1^2+y_1^2}} & \frac{y_1}{\sqrt{x_1^2+y_1^2}} & 0 \\
 -\frac{y_1}{\sqrt{x_1^2+y_1^2}} & \frac{x_1}{\sqrt{x_1^2+y_1^2}} & 0 \\
0 & 0 & 1
\end{array} 
\right),\\
\lambda_i &=&\frac{1}{(R(P_i-P_0))_x}-1.
\end{array}
\end{equation}
where $f=\frac{-y_1 x_2 +x_1 y_2}{\sqrt{x_1^2+y_1^2}}$, $g=\frac{z_2 (x_1^2+y_1^2)-z_1 (x_1 x_2+y_1 y_2)}{\sqrt{x_1^2+y_1^2}\sqrt{x_1^2+y_1^2+z_1^2}}$, $(x_1, y_1,z_1)^T=P_1-P_0 $ and $(x_2,y_2,z_2)^T=P_2-P_0$.
These group parameters define a moving frame (MF).
Replacing the moving frame 
into the transformation equations, we get:
\begin{eqnarray*}
\left. \bar{P}_0 \right|_{MF} &= &(0,0,0)^T ,\\
\left. \bar{P}_1 \right|_{MF} &=&(1,0,0)^T ,\\
\left. \bar{P}_2 \right|_{MF} &=& \left(
\begin{array}{c}
1 \\
\frac{ f\sqrt{x_1^2+y_1^2+z_1^2}(x_1y_2-x_2y_1)+g[z_2(x_1^2+y_1^2)-z_1(x_1x_2+y_1y_2)] }{ (x_1x_2+y_1y_2+z_1z_2) \sqrt{x_1^2+y_1^2}\sqrt{f^2+g^2} } \\
0\\
\end{array}\right),\\
\left. \bar{P}_i \right|_{MF} &=& \left(
\begin{array}{c}
1 \\
\frac{ f\sqrt{x_1^2+y_1^2+z_1^2}(x_1y_i-x_iy_1) +g[z_i(x_1^2+y_1^2)-z_1(x_1x_i+y_1y_i)] }{ (x_1x_i+y_1y_i+z_1z_i) \sqrt{x_1^2+y_1^2}\sqrt{f^2+g^2} }\\
\frac{g\sqrt{x_1^2+y_1^2+z_1^2}(x_iy_1-x_1y_i) +f[z_i(x_1^2+y_1^2)-z_1(x_1x_i+y_1y_i)] }{ (x_1x_i+y_1y_i+z_1z_i) \sqrt{x_1^2+y_1^2}\sqrt{f^2+g^2} }\\
\end{array}\right).\\
\end{eqnarray*}
for all $i=3,\ldots, n$, where $(x_i, y_i, z_i)=P_i-P_0$. 
Each  component of these vectors is an invariant of the group action.

Let us try to understand the geometric meaning of these expression.
Observe that $\sqrt{f^2+g^2}=\frac{\|(x_1,y_1,z_1)\times (x_2,y_2,z_2)\|}{\sqrt{x_1^2+y_1^2+z_1^2}}$. After a few manipulations, we can rewrite the above system as:
\begin{eqnarray*}
\left. \bar{P}_0 \right|_{MF} &= &(0,0,0)^T ,\\
\left.\bar{P}_1 \right|_{MF}&=&(1,0,0)^T\\
\left.\bar{P}_2 \right|_{MF} &=& \left(
\begin{array}{c}
1 \\
\frac{\| (x_1,y_1,z_1)\times (x_2,y_2,z_2) \| }{(x_1,y_1,z_1) \cdot (x_2,y_2,z_2)} \\
0 \\
\end{array}\right),\\
\left.\bar{P}_i \right|_{MF} &=& \left(
\begin{array}{c}
1 \\
\frac{[( x_1,y_1,z_1)\times (x_i,y_i,z_i)] \cdot [(x_1,y_1,z_1)\times (x_2,y_2,z_2)] }{[(x_1,y_1,z_1) \cdot (x_i,y_i,z_i)] \: \| (x_1,y_1,z_1) \times (x_2,y_2,z_2) \|}\\
\frac{  (x_i, y_i, z_i) \cdot [(x_2, y_2, z_2) \times (x_1, y_1, z_1)] \| (x_1,y_1,z_1) \|}{ [(x_1,y_1,z_1) \cdot (x_i,y_i,z_i) ] \: \| (x_1,y_1,z_1) \times (x_2,y_2,z_2) \|} \\
\end{array}\right).\\
\end{eqnarray*}

It now becomes clearer that the components of $\left.\bar{P}_2 \right|_{MF}$ and $\left.\bar{P}_i \right|_{MF}$ are sine or cosine of angles between the directions spanned by $\overline {P_1 P_0}$, $\overline {P_2 P_0}$, $\overline {P_i P_0}$ and the directions orthogonal to them.
These are clearly invariant by translation, rotation, and motion along the projection lines. 
As a fundamental set, we simply pick the only $2n-3$ non-constant invariants:
\begin{eqnarray*}
I_2 &=& \frac{\|(x_1,y_1,z_1)\times (x_2,y_2,z_2) \| }{(x_1,y_1,z_1) \cdot (x_2,y_2,z_2)} \\
I_i &=& \frac{[( x_1,y_1,z_1)\times (x_i,y_i,z_i)] \cdot [(x_1,y_1,z_1)\times (x_2,y_2,z_2)] }{[(x_1,y_1,z_1)\cdot (x_i,y_i,z_i)] \: \|(x_1,y_1,z_1) \times (x_2,y_2,z_2) \|}\\
J_i &=& \frac{ (x_i, y_i, z_i)\cdot[(x_2, y_2, z_2)\times(x_1, y_1, z_1)]  \|(x_1,y_1,z_1)\| }{[(x_1,y_1,z_1)\cdot (x_i,y_i,z_i)] \: \|(x_1,y_1,z_1) \times (x_2,y_2,z_2)\|} \\
\end{eqnarray*}
for $i=3, \ldots, n$.

Each picture taken defines a point  in ${\mathbb R}^3\times ({\mathbb R}^3)^{\times (n)}$ and therefore determines
an orbit of our group action.
Each orbit is characterized by the set of $2n-3$ equations given by the invariants. More precisely, indexing the pictures with the discrete parameter $\tau =1,\ldots, t $, 
we have
\begin{eqnarray*}
I_i(P_0^\tau,P_1,\ldots, P_n)&=&\alpha_i^\tau, \text{ for }i=2,\ldots,n, \\
J_j(P_0^\tau,P_1,\ldots, P_n)&=&\beta_j^\tau, \text{ for }j=3,\ldots,n. \\
\end{eqnarray*}
for appropriate constants $\alpha_i^\tau$'s and $\beta_j^\tau$'s. 
These constants are  prescribed by the pictures: 
since the picture-camera system itself belongs to the orbits, we have
\begin{eqnarray*}
\alpha_i^\tau &=&I_i(\tilde{p}_0^\tau,\tilde{p}_1^\tau,\ldots ,\tilde{p}_n^\tau)\\
\beta_j^\tau &=&J_j(\tilde{p}_0^\tau,\tilde{p}_1^\tau,\ldots ,\tilde{p}_n^\tau)\\ 
\end{eqnarray*}

We are interested in solving the equations
\begin{eqnarray*}
I_i({\mathfrak P}_0^\tau,{\mathcal O}_1,\ldots,{\mathcal O}_n )&=&\alpha_i^\tau, \text{ for }i=2,\ldots,n \\
J_j^\tau({\mathfrak P}_0^\tau,{\mathcal O}_1,\ldots,{\mathcal O}_n )&=&\beta_j^\tau, \text{ for }j=3,\ldots,n. \\
\end{eqnarray*}
for $\tau=1,\ldots, t$.
For 

We have $(2n-3)t$ (non-linear) equations with $3n+3t$ unknowns, the solution of which is determined up to a rotation and translation of the 3D camera-object system as a whole, which can can fix arbitrarily, thus eliminating six variables\footnote{However, we should keep in mind that the choice of these variables will affect the numerical resolution\cite{kanatani00b}.}.
For $n>3$ and $t\geq \frac{3n-6}{2n-6}$, the number of equation is greater than the number of unknown so we can try to solve them.

Experiments with real video images have been performed (see Fig.\ref{fig:reconstruct} using a sequential non-linear optimization technique based on the Levenberg-Marquardt algorithm \cite{press93}. 
The feature points used in the pictures are endpoints of lines and rectangles associated from one image to the next with a tracking procedure \cite{bazin00b}. 
The reconstructed 3D object is valid in any view, even if the bottom and left side elements are not perfectly replaced, due to the tracking noise. The computations take only a few minutes.

\Figure{
\begin{tabular}{cccc}
\includegraphics[height=2.5cm]{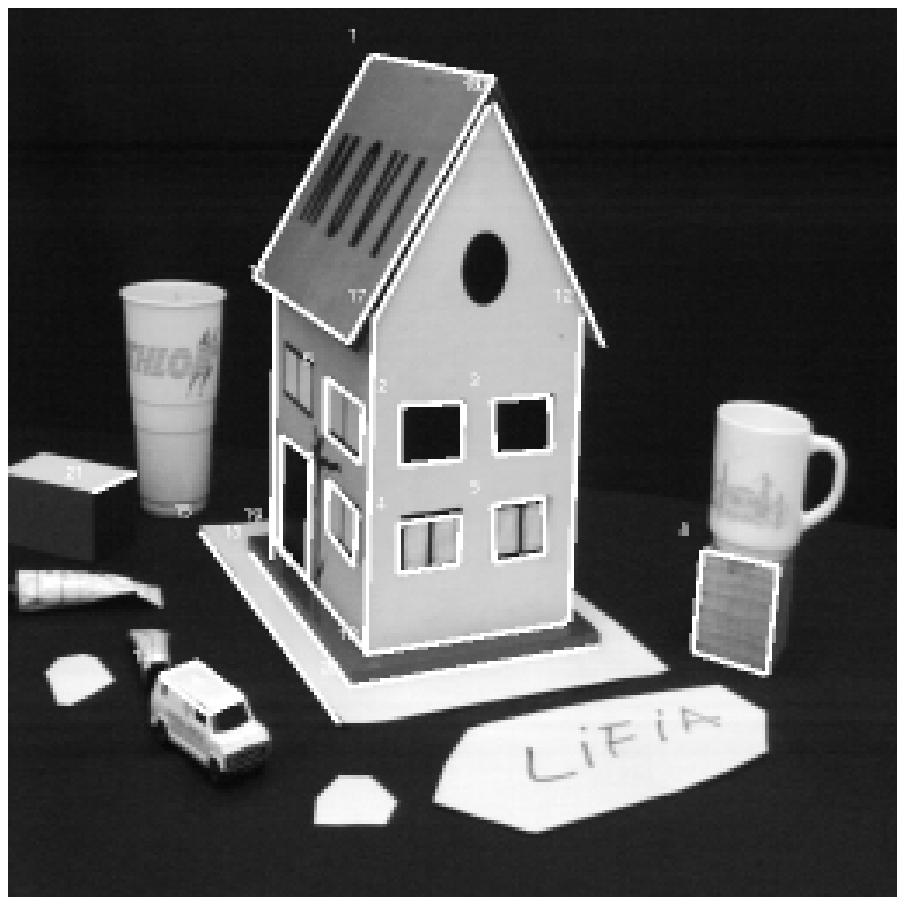} &
\includegraphics[height=2.75cm]{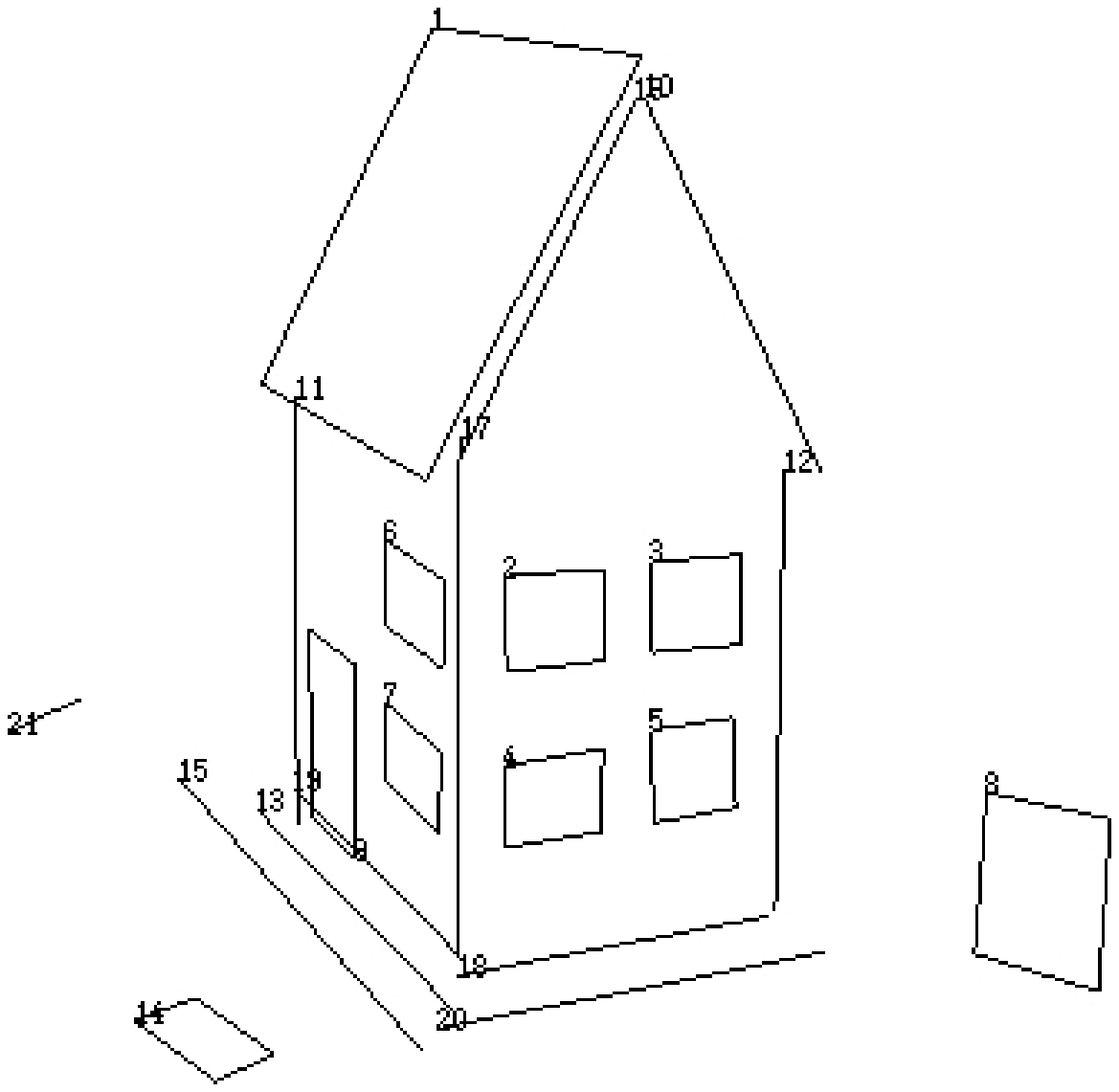} &
\includegraphics[height=2.75cm]{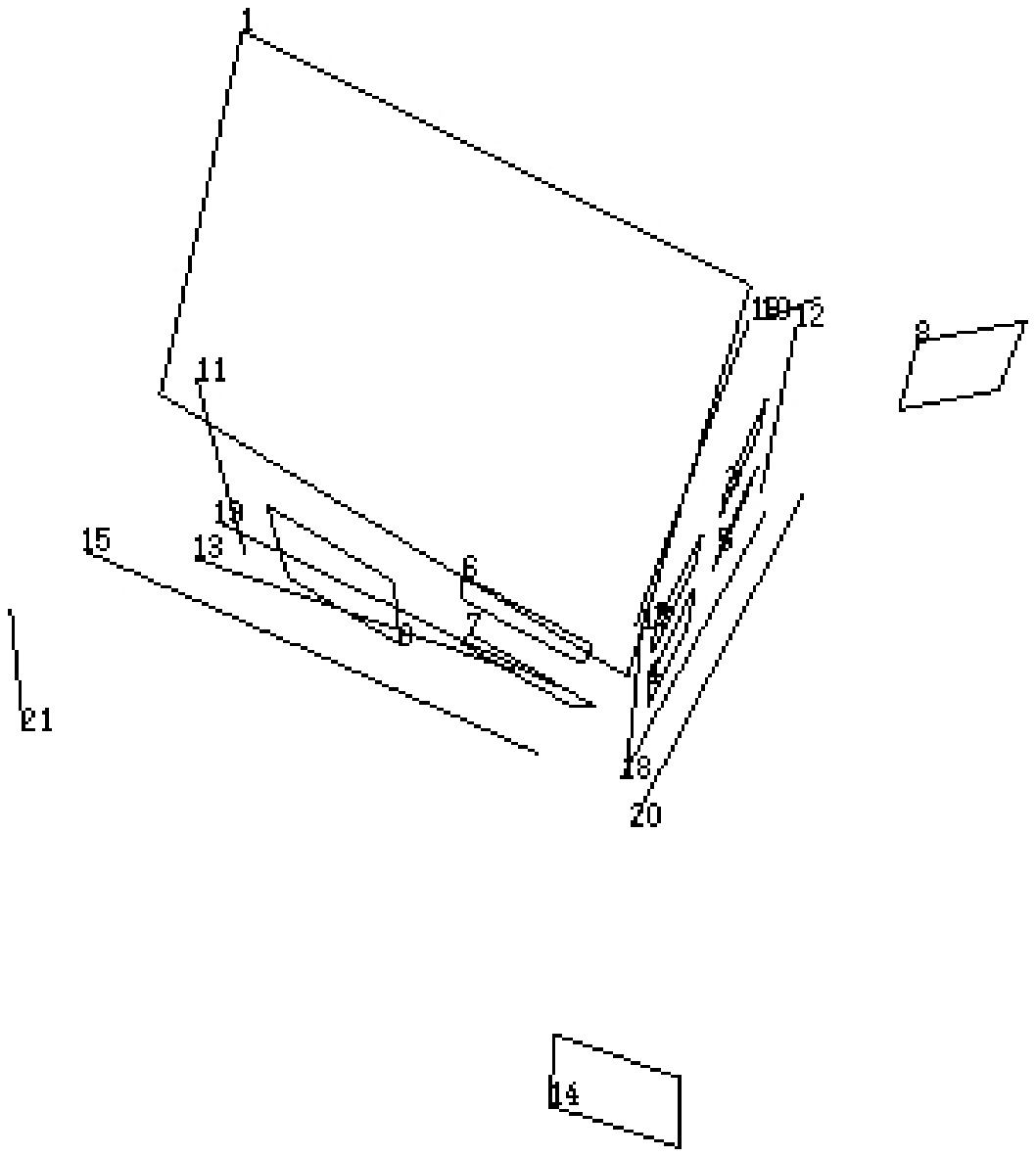} &
\includegraphics[height=2.5cm]{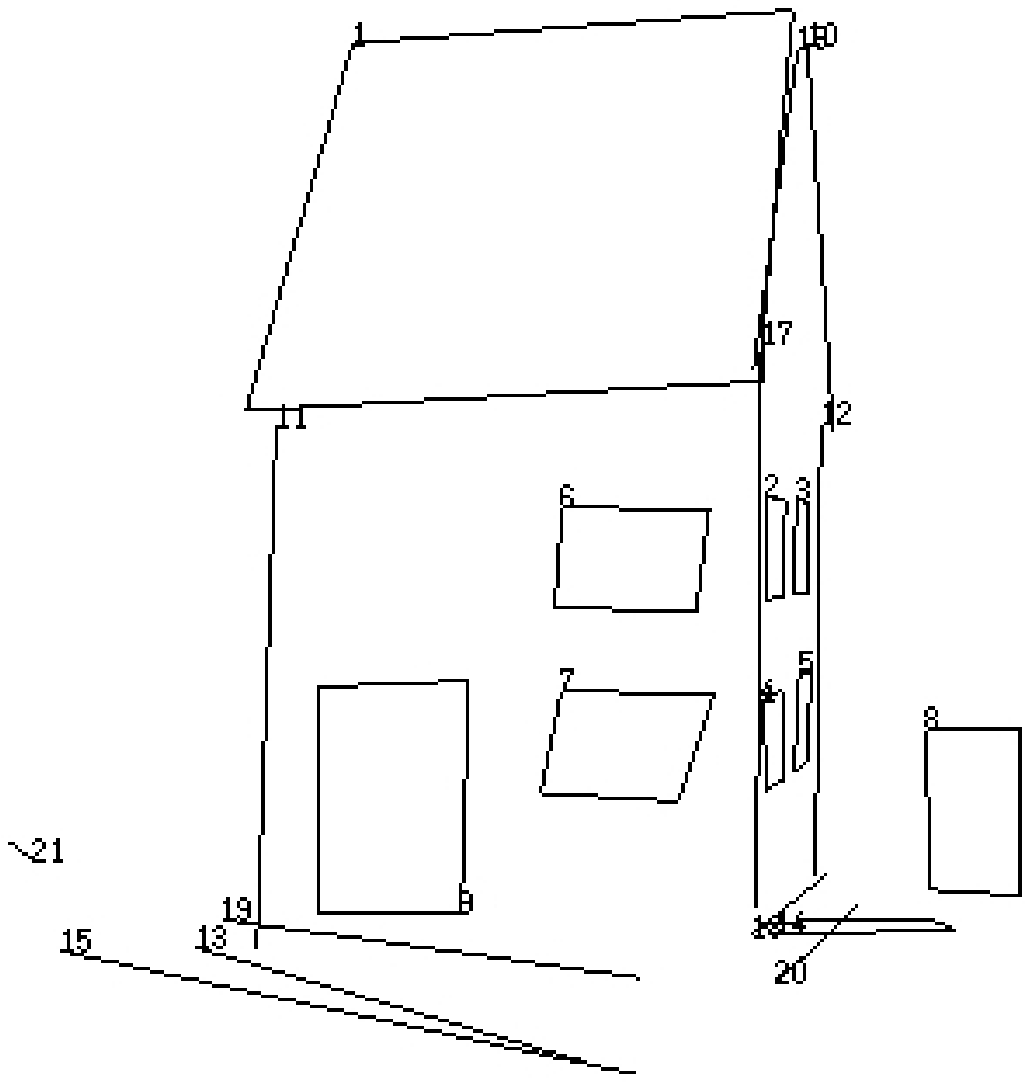} \\
a & b & c & d \\
\end{tabular}
}{A reconstruction example: a) the first of six images, with line and rectangle features, b) a similar view of the 3D reconstructed features, c) a top view, d) a side view of the reconstruction.}{fig:reconstruct}

Observe that our camera-system does not take into account the angle of the camera; the orientation of the image plane was not used in our description of the camera-picture system, Our invariants are invariant under a rotation of the image plane and so we cannot use them to recover the camera orientation. 
But this has its advantages, as illustrated by the following lemma.

\begin{lemma}\mylabel{pilou's corollary version 1}
The motion of the camera between two pictures is a pure rotation (i.e.~a rotation around the center of projection $P_0$) 
if and only if the values of  the invariants $\{ I_i, J_j | i=2,\ldots, n, j=3,\ldots, n\}$ 
evaluated on any corresponding points in the two views are equal.
\end{lemma} 
\begin{proof}
Invariance of our invariants under pure rotations is obvious from the construction of the invariants.

To prove that  equality of our invariants evaluated on all corresponding points guarantees that the camera motion is a pure rotation, 
observe that  the first invariant $I_2$ is the tangent of the angle between the lines $\overline{{\mathfrak P}_0 {\mathcal O}_1}$ and $\overline{{\mathfrak P}_0 {\mathcal O}_2}$. 
Its value remains constant for fixed ${\mathcal O}_1,{\mathcal O}_2$ only if ${\mathfrak P}_0$ moves along a circle around the $\overline{{\mathcal O}_1 {\mathcal O}_2}$ axis.
This holds for all possible choices of  
${\mathcal O}_1$ and ${\mathcal O}_2$ 
so the camera center must lie somewhere on the intersection of a set of circles, which can be taken to intersect at merely one point to guarantee that the camera center does not move.
\end{proof}

\section{Variations on the theme}

\subsection{Taking the angle of the camera plane into account}

If we want to recover the camera rotation, we need to modify the system to introduce orientation information. 
Let $P_L$ denote the lower left corner of the camera plane (embedded in ${\mathbb R}^3$) and  $P^L$ the upper left corner of the camera plane.
These two points are sufficient to uniquely define the orientation of the camera plane. 
Consider the following Lie group action on $\{(P_0, P_L,P^L,P_1,\ldots ,P_n)\in ({\mathbb R}^3)^{\times (3)}\times ({\mathbb R}^3)^{\times (n)} \}$:

\begin{eqnarray*}
\bar{P}_0 &= & R P_0 +T \\
\bar{P}_L &= & R P_L +T \\
\bar{P}^L &= & R P^L +T \\
\bar{P}_i &= & R (P_i+\lambda_i (P_i-P_0))+T, \text{ for }i=1,\ldots, n, \\
\end{eqnarray*}
with $R\in SO(3)$ a rotation, $T\in {\mathbb R}^3$ a translation and $\lambda_i\in {\mathbb R}$, a factor of depth, for $i=1,\ldots, n$.
Again, this defines  a finite dimensional Lie group action, more precisely the action of a $(6+n)$-dimensional Lie group on a $(3n+9)$-dimensional manifold.

Assuming that the points are distinct, this group action is regular and
the orbits are $6+n$-dimensional, for any $n$.
Therefore by Theorem \ref{fundamental theorem}, there are $2n+3$ 
fundamental invariants.

We set 
\begin{eqnarray*}
\bar{P}_0 &= &(0,0,0)^T ,\\
\bar{P}_L \cdot (0,1,0) &=&0,\\
\bar{P}_L \cdot (0,0,1) &=&0,\\
\bar{P}^L \cdot (0,0,1) &=&0,\\
\text{and } \bar{P}_i \cdot (1,0,0) &=&1, \text{ for all }i=1,\ldots,n.
\end{eqnarray*}

For simplicity, we recycle the variables used for the moving frame in the previous case. Solving for the group parameters, we obtain a very similar result, with $T$, $R$ and $\lambda_i$ defined as in eq. (\ref{eqn:first}), except that $(x_1, y_1,z_1)^T$ and $(x_2,y_2,z_2)^T$ are replaced by $(u_1,v_1,w_1)^T=P_L-P_0 $ and $(u_2,v_2,w_2)=P^L-P_0$ respectively.
These group parameters define a moving frame for the new transformation group.
Replacing the moving frame 
into the transformation equations, we get a similar yet different result:
\begin{eqnarray*}
\left. \bar{P}_0 \right|_{MF} &= &(0,0,0)^T ,\\
\left. \bar{P}_L \right|_{MF} &=&(\| P_L-P_0 \|,0,0)^T ,\\
\left. \bar{P}^L \right|_{MF} &=& \left(
\begin{array}{c}
\frac{(u_2,v_2,w_2) \cdot(u_1,v_1,w_1) }{||(u_1,v_1,w_1)|| } \\
\frac{||(u_1,v_1,w_1)\times (u_2,v_2,w_2) || }{||(u_1,v_1,w_1)||}\\
0
\end{array}\right),\\
\left.\bar{P}_i \right|_{MF} &=& \left(
\begin{array}{c}
1 \\
\frac{[( u_1,v_1,w_1)\times (x_i,y_i,z_i)] \cdot [(u_1,v_1,w_1)\times (u_2,v_2,w_2)] }{[(u_1,v_1,w_1) \cdot (x_i,y_i,z_i)] \: \| (u_1,v_1,w_1) \times (u_2,v_2,w_2) \|}\\
\frac{  (x_i, y_i, z_i) \cdot [(u_2, v_2, w_2) \times (u_1, v_1, w_1)] \| (u_1,v_1,w_1) \|}{ [(u_1,v_1,w_1) \cdot (x_i,y_i,z_i) ] \: \| (u_1,v_1,w_1) \times (u_2,v_2,w_2) \|} \\
\end{array}\right).\\
\end{eqnarray*}
for all $i=3,\ldots, n$, where $(x_i, y_i, z_i)=P_i-P_0$. 

The new set of fundamental invariants for this group action is:
\begin{eqnarray*}
I_L &=& \| P_L-P_0 \| \\
I_0 &=& \frac{(u_2,v_2,w_2) \cdot(u_1,v_1,w_1) }{||(u_1,v_1,w_1)|| } \\
J_0 &=&
\frac{||(u_1,v_1,w_1)\times (u_2,v_2,w_2) || }{ ||(u_1,v_1,w_1)||}   \\
I_i &=& \frac{[( u_1,v_1,w_1)\times (x_i,y_i,z_i)] \cdot [(u_1,v_1,w_1)\times (u_2,v_2,w_2)] }{[(u_1,v_1,w_1)\cdot (x_i,y_i,z_i)] \: \|(u_1,v_1,w_1) \times (u_2,v_2,w_2) \|}\\
J_i &=& \frac{ (x_i, y_i, z_i)\cdot[(u_2, v_2, w_2)\times(u_1, v_1, w_1)]  \|(u_1,v_1,w_1)\| }{[(u_1,v_1,w_1)\cdot (x_i,y_i,z_i)] \: \|(u_1,v_1,w_1) \times (u_2,v_2,w_2)\|} \\
\end{eqnarray*}
for $i=1, \ldots, n$.

Each picture taken defines a point  in $({\mathbb R}^3)^{\times (3)}\times ({\mathbb R}^3)^{\times (n)}$ and therefore determines
an orbit of our group action.
Each orbit is characterized by the set of $2n+3$ equations given by the invariants. Given $t$ pictures, we have to solve a system of $(2n+3)t$ equations with $3n+9t$ unknowns, 6 of which can be fixed arbitrarily. 
Again for $t$ and $n$ big enough ($n>3$ is necessary, $t \geq 3$ for $n=4$, etc.),  
the number of equation is superior to the number of variables. We can search for a  solution within the same optimization framework used in the previous case.
The values of ${\mathfrak P}_0^\tau$, ${\mathfrak P}_L^\tau$ and ${\mathfrak P}^{L\tau}$ must be included in the solution. We thus obtain both  camera position and orientation for each picture.

\subsection{Letting the focal distance vary}

It is slightly less trivial to set up the group transformation equations in the case were the focal length is allowed to vary from one image to the next.  Consider $P_M$, the closest point to $P_0$ on the image plane, i.e.~the embedding of the middle point of the picture. Changing the focal distance corresponds to transforming $P_M$ into a point $P'_M$ with a real parameter $\alpha$ according to the rule
\[P'_M=P_M+\alpha (P_M-P_0).
\]
The induced action on the $P_i$'s can be taken as a rotation about the center of camera $P_0$ which preserves the distance to the camera center. 
More precisely, 
each $P_i$ is moved to a new point $P'_i$ in such a way that $\| P'_i- P_0 \|=\|P_i- P_0 \|$ and that its transformed picture point is $\tilde{p}'_i = \tilde{p}_i + \alpha (P_M-P_0)$.
Since the picture point is given by $p_i = P_0 + \frac{P_i-P_0}{(P_i-P_0)\cdot (P_M-P_0)}$, we find that
$$
P'_i = P_0 + \|P_i- P_0 \|\quad \frac{\frac{P_i-P_0}{(P_i-P_0)\cdot (P_M-P_0)}+ \alpha (P_M- P_0)}{\|\frac{P_i-P_0}{(P_i-P_0)\cdot (P_M-P_0)}+ \alpha (P_M- P_0) \|}.
$$
Combining with translations of the $P_i$'s along the line $P_iP_0$ together with rotations and translations of the line arrangement as a whole, 
we get the following $(n+7)$-dimensional Lie group action:
\begin{eqnarray*}
\bar{P}_0 &=& R P_0+T,\\
\bar{P}_M &=& R\left( P_M+\alpha (P_0-P_M)\right)+T,\\
\bar{P}_i &=& R\left(P_0 + (1+\lambda_i)\|P_i- P_0 \|\frac{\frac{P_i-P_0}{(P_i-P_0)\cdot (P_M-P_0)}+ \alpha (P_M- P_0)}{\|\frac{P_i-P_0}{(P_i-P_0)\cdot (P_M-P_0)}+ \alpha (P_M- P_0) \|}  \right)+T.
\end{eqnarray*}
Observe that the change of focal length parameterized by $\alpha$ commutes with the action of each $\lambda_i$ on  $P_i$, because it preserves the norm $\|P_i- P_0 \|$.

We apply the previous moving frame normalization technique by setting:
\begin{eqnarray*}
\bar{P}_0 &= &(0,0,0)^T ,\\
\bar{P}_M &= &(1,0,0)^T ,\\
\bar{P}_1 \cdot (0,0,1) &=&0,\\
\text{and } \bar{P}_i \cdot (1,0,0) &=&1, \text{ for all }i=1,\ldots,n.
\end{eqnarray*}

The corresponding group parameters are similar to eq. (\ref{eqn:first}) for $R$ and $T$, except that 
 $(x_1, y_1,z_1)^T$ and $(x_2,y_2,z_2)^T$ must be replaced by $(u_1,v_1,w_1)^T=P_M-P_0 $ and $(u_2,v_2,w_2)^T=P_1-P_0$. The other parameters are:
\begin{eqnarray*}
\alpha &= & \frac{1}{\|P_M-P_0\|}-1, \\
\lambda_i &= & \frac{\|\frac{P_i-P_0}{(P_i-P_0)\cdot (P_M-P_0)}+ \alpha (P_M- P_0) \|}{\|P_i-P_0\|\|P_M-P_0\|+\frac{\|P_i-P_0\|}{\|P_M-P_0\|}}-1, \text{ for all }i=1,\ldots,n.
\end{eqnarray*}
Replacing these group parameters into the equations for the $\bar{P}_i$'s, we obtain the following complete fundamental set of invariants:
{\small
\begin{eqnarray*}
I_1 &=&\frac{\|(P_M-P_0)\times (P_1-P_0)\|}{(P_1-P_0)\cdot (P_M-P_0) \left(1+\|P_M-P_0\|-\|P_M-P_0\|^2 \right)}, \\ 
I_i &=&\frac{(P_M-P_0)\times (P_i-P_0)\cdot (P_M-P_0)\times (P_1-P_0)}{\|(P_M-P_0)\times (P_1-P_0) \| (P_i-P_0)\cdot (P_M-P_0) \left(1+\|P_M-P_0\|-\|P_M-P_0\|^2 \right)}, \\ 
J_i &=&\frac{ (P_i-P_0)\cdot \left[ (P_1-P_0)\times (P_M-P_0) \right] \| P_M-P_0 \| }{ \|(P_M-P_0)\times (P_1-P_0) \|  (P_i-P_0)\cdot (P_M-P_0) \left( 1+\| P_M-P_0\|-\|P_M-P_0\|^2 \right) },  
\end{eqnarray*}
}
for $i=2,\ldots, n$.

\subsection*{Acknowledgments}
Both authors would like to thank David Cooper for encouragement and stimulating discussions.


\end{document}